\documentclass{article}
\usepackage{ijcai22}

\usepackage{times}
\usepackage{soul}
\usepackage{url}
\usepackage[hidelinks]{hyperref}
\usepackage[utf8]{inputenc}
\usepackage[small]{caption}
\usepackage{graphicx}
\usepackage{amsmath}
\usepackage{amsthm}
\usepackage{booktabs}
\usepackage{algorithm}
\usepackage{algorithmic}
\usepackage{stmaryrd}

\usepackage{graphicx}
\usepackage{latexsym}

\usepackage{amsthm}

\usepackage{amsmath} \usepackage{tikz-cd} \usepackage {tikz}
\usetikzlibrary{positioning,graphs,calc,decorations.pathmorphing,shapes,arrows.meta,arrows,shapes.misc}
\usetikzlibrary{fit}
\usetikzlibrary{arrows}
\usepackage{comment}

\newcommand{\defi}{\doteq}

\newtheorem{definition}{Definition}

\title{Behaviour Explanation via Causal Analysis of Mental States: \\A Preliminary Report}
\author{
    Shakil M. Khan
    \affiliations
    University of Regina, Saskatchewan, Canada
    \emails
    shakil.khan@uregina.ca
}

\begin{document}

\maketitle

\begin{abstract}
Inspired by a novel action-theoretic formalization of actual cause, Khan and Lesp\'{e}rance (2021) recently proposed a first account of causal knowledge that supports epistemic effects, models causal knowledge dynamics, and allows sensing actions to be causes of observed effects. To date, no other study has looked specifically at these issues. But their formalization is not sufficiently expressive enough to model explanations via causal analysis of mental states as it ignores a crucial aspect of theory of mind, namely motivations. In this paper, we build on their work to support causal reasoning about conative effects. In our framework, one can reason about causes of motivational states, and we allow motivation-altering actions to be causes of observed effects. We illustrate that this formalization along with a model of goal recognition can be utilized to explain agent behaviour in communicative multiagent contexts. 
\end{abstract}


%
\section{Introduction}
\emph{Actual causality} is a long-standing philosophical problem that is fundamental to the task of reasoning about and explaining observations. Given a narrative or history of events and an observed effect, solving this problem involves finding the events or actions from this history that are responsible for producing this effect, i.e.\ those that caused the effect. Also known as {\em token-level} causality, this problem is different from \emph{general} or {\em type-level} causality, where the task is to discover universal causal mechanisms. Actual causality plays a significant role in reasoning about agents. For instance, causal reasoning can be used to explain the behaviour of a group of agents, e.g.\ via causal analysis of the mental states produced by this behaviour. These mental states may include beliefs and goals of the agents whose actions are the cause of the observed behaviour as well as those of others' (see Sec.\ 5 for an example).
\par
Pearl \shortcite{Pearl98,Pearl00} was a pioneer in computational enquiry into actual causality. This line of research was later continued by Halpern \shortcite{Halpern00}, Halpern and Pearl \shortcite{HalpernP05}, and others \cite{EiterL02,Hopkins05,HopkinsP07,Halpern15,Halpern16}. This ``HP approach'' is based on the concept of structural equations \cite{Simon77}. HP follows the Humean counterfactual definition of causation, which states that ``an outcome $B$ is caused by an event $A$'' is the same as saying that ``had $A$ never occurred, $B$ never had existed''. This definition suffers from the problem of preemption\footnote{Preemption happens when two competing events try to achieve the same effect, and the latter of these fails to do so, as the earlier one has already achieved the effect.}: it could be the case that in the absence of event $A$, $B$ would still have occurred due to another event, which in the original trace was preempted by $A$. HP address this by performing counterfactual analysis only under carefully selected contingencies, which suspend some subset of the model's mechanisms. While their inspirational early work was shown to be useful for some practical applications,  
their approach based on Structural Equations Models (SEM) has been criticized for its limited expressiveness \cite{Hopkins05,HopkinsP07,GlymourDGERSSTZ10}, and researchers have attempted to expand SEM with additional features, e.g.\ \cite{Leitner-FischerL13}. Note that despite recently reported progresses (e.g.\ \cite{HalpernPeters22}), many of these expressive limitations remain. Also, while there has been much work on actual causality, the vast majority of the work in this area has focused on defining causes from an objective standpoint.
\par
In recent years, researchers have become increasingly interested in studying causation from the perspective of agents. Among other things, this is useful for defining important concepts such as responsibility and blame. Inspired by a novel action-theoretic formalization of actual causation \cite{BatusovS18}, Khan and Lesp\'{e}rance \shortcite{KL21} (KL, henceforth) recently proposed a first account of causal knowledge that supports epistemic effects, models causal knowledge dynamics, and allows sensing actions to be causes of observed effects. To date, no other study has looked specifically at these issues. But their formalization is not sufficiently expressive enough to model explanations via causal analysis of mental states as it ignores a crucial aspect of theory of mind, namely motivations. In this paper, we build on their work to support causal reasoning about conative effects. In our framework, one can reason about causes of motivational states, and we allow motivation-altering actions to be causes of observed effects. We illustrate that this formalization along with a model of goal recognition can be utilized to explain agent behaviour. 
\par
Our contribution in this paper is three-fold. First, we show how causal reasoning about goals/intentions can be modeled. Secondly, using an example, we illustrate how this formalization along with a model of goal recognition can be used to explain agent behaviour in communicative multiagent contexts. The generated explanations include both direct causal explanations as well as higher-order and more useful indirect explanations. The latter is grounded in (multiagent) theory of mind-based causal reasoning. Finally, while doing this, we extend a previously proposed account of goal change to deal with the \emph{request} communicative action.
\section{Action and Knowledge}
\textbf{The Situation Calculus}\indent
Our base framework for modeling causal reasoning is the situation calculus (SC) \cite{McCarthyH69} as formalized in \cite{Reiter01}. 
Here, a possible state of the domain is represented by a situation. The initial state is denoted by $S_0$. There is a distinguished binary function symbol $do$ where 
$do(a,s)$ denotes the successor situation to $s$ resulting from performing the action $a$. 
Thus the situations can be viewed forming a tree, where the root of the tree is an initial situation and the arcs represent actions. 
As usual, a relational/functional fluent takes a situation term as its last argument. There is a special predicate $\mathit{Poss}(a,s)$ used to state that action $a$ is executable in situation $s$. 
We will use the abbreviation $do([\alpha_1,\cdots,\alpha_n],S_0)$ to represent the situation obtained by consecutively performing $\alpha_1,\cdots,\alpha_n$ starting from $S_0$. 
Also, the notation $s\sqsubset s'$ means that situation $s'$ can be reached from situation $s$ by executing a sequence of actions. $s\sqsubseteq s'$ is an abbreviation of $s\sqsubset s'\vee s= s'.$ 
$s<s'$ is an abbreviation of $s\sqsubset s'\wedge \mathit{executable}(s'),$ where 
$\mathit{executable}(s)$ is defined as $\forall a',s'.\;do(a',s')\sqsubseteq s\supset \mathit{Poss}(a',s'),$ i.e.\ every action performed in reaching situation s was possible in the situation in which it occurred. 
$s\leq s'$ is an abbreviation of $s<s'\vee s= s'.$
\par
Our framework uses an action theory $\mathcal{D}$ that includes the following set of axioms:\footnote{We will be quantifying 
over formulae, and thus assume that $\mathcal{D}$ includes axioms for encoding of formulae as first order terms, as in \cite{ShapLesLev07}.} 
(1) action precondition axioms (APA), one per action $a$ characterizing $\mathit{Poss}(a,s)$,
(2) successor state axioms (SSA), one per fluent, that succinctly encode both effect and frame axioms and specify 
exactly when the fluent changes, 
(3) initial state axioms describing what is true initially, 
(4) unique name axioms for actions, and    
(5) domain-independent foundational axioms describing the structure of situations \cite{LevPirRei98}. 
%
%
\newline\noindent\textbf{Knowledge in the Situation Calculus}\indent
Following \cite{Moore85,SchLev03}, we model knowledge using a possible worlds account adapted to the SC. 
There can now be multiple initial situations. $\mathit{Init}(s)$ means that $s$ is an initial situation. 
The actual initial state is denoted by $S_0$. 
$K(d,s',s)$ is used to denote that in situation $s$, the agent $d$ thinks that it could be in situation $s'$. 
Using $K$, the knowledge of an agent $d$ is defined as:\footnote{$\Phi$ can contain a placeholder {\em now} 
in the place of the situation terms. We often suppress {\em now} when the intent is clear from the context. 
Also, $\Phi[s]$ denotes the formula obtained by restoring the situation argument $s$ into all fluents in $\Phi$. 
}  
$\mathit{Know}(d,\Phi,s)\defi\forall s'.\;K(d,s',s)\supset \Phi[s']$, i.e.\ the agent $d$ knows $\Phi$ in $s$ if $\Phi$ holds in 
all of its $K$-accessible situations in $s$. We also use the abbreviations $\mathit{Kwhether}(d,\Phi,s)\defi \mathit{Know}(d,\Phi,s)\vee \mathit{Know}(d,\neg\Phi,s),$ i.e.\ $d$ knows whether $\Phi$ holds in $s$ 
and $\mathit{Kref}(d,\theta,s)\defi\exists t.\;\mathit{Know}(d,\theta=t,s)$, i.e.\ it knows who/what $\theta$ refers to in $s$. 
$K$ is constrained to be reflexive 
and Euclidean (and thus transitive) in the 
initial situation to capture the fact that the agent's knowledge is true, and that it has positive and negative introspection. 
\par
In our framework, the dynamics of knowledge is specified using a SSA for $K$ that supports knowledge 
expansion as a result of sensing actions as well as communication actions. The information provided by a binary sensing action is specified using the predicate $SF(a, s)$.  
%
Similarly for non-binary sensing actions, the term $\mathit{sff}(a, s)$ is used to denote the sensing value returned by the action. 
%
These are specified using {\em sensed fluent axioms}; see \cite{Lev96} for details.
Shapiro et al. \shortcite{ShapiroLL97} and later Lesp\'erance \shortcite{Les02} extended the SSA for $K$ to support variants of the `inform' communicative action. We will adopt the variant proposed in KL \shortcite{KL05}. The preconditions of $\mathit{inform}(\mathit{inf},\mathit{agt},\Phi)$, which can be used by $\mathit{inf}$ to inform $\mathit{agt}$ that $\Phi$, are as follows:
\begin{eqnarray*}
&&\mathit{Poss}(\mathit{inform}(\mathit{inf},\mathit{agt},\Phi),s)\equiv\mathit{Know}(\mathit{inf},\Phi,s)\\
&&\hspace{10 mm}\mbox{}\wedge\neg\mathit{Know}(\mathit{inf},\mathit{Know}(\mathit{agt},\Phi,\mathit{now}),s).
\end{eqnarray*}
We assume that its effects has been specified as in KL \shortcite{KL05}. 
As shown in \cite{SchLev03}, the constraints on $K$ then continue to hold after any sequence of actions since they are preserved by the SSA for $K$. A similar result can be shown for the KL \shortcite{KL05} variant of the SSA for $K$.
%
%
\par
Thus to model knowledge, we will use a theory that is similar to before, but with modified foundational axioms to allow for multiple initial epistemic states. Also, action preconditions can now include knowledge preconditions and initial state axioms can now include axioms describing the epistemic states of the agents. Finally, the aforementioned axioms for $K$ and $\mathit{inform}$ are included. See \cite{Reiter01} and \cite{KL05} for details of these. 
Note that like \cite{SchLev03}, we assume that actions are fully observable (even if their effects are not). This can be generalized as in \cite{BacchusHL99}.
%
\newline\noindent\textbf{Paths in the Situation Calculus}\indent
Following KL \shortcite{KL16}, we will formalize the sort of \emph{paths} in the SC.  A path is essentially an infinite sequence of situations, where each situation along the path can be reached by performing some executable action in the preceding 
situation. We will use $\mathit{Starts}(p,s)$ to denote that $s$ is the earliest situation on path $p$ and $\mathit{OnPath}(p,s)$ to denote that $s$ is on $p$. $\mathit{Suffix}(p',p,s)$ means that path $p'$ that starts with situation $s$ is a suffix of 
$p$. 
KL \shortcite{KL15} showed how one can interpret arbitrary CTL$^*$ formulae within SC with paths. We assume that our theory $\mathcal{D}$ includes the axiomatization for paths.
\par
We will use uppercase and lowercase Greek letters for state formulae (i.e.\ situation-suppressed SC formulae) and path formulae, resp. 
These are inductively defined as follows:

\vspace{-4 mm}\begin{small}
\begin{eqnarray*}
&&\hspace{0 mm}\Phi::=P(\vec{x})\mid A \phi\mid\Phi\wedge\Phi\mid\neg\Phi\mid\forall x.\;\Phi\\
&&\hspace{0 mm}\phi::=\Phi\mid\phi\wedge\phi\mid\neg\phi\mid\forall x.\;\phi\mid\bigcirc\phi\mid\phi\;\mathcal{U}\;\phi
\end{eqnarray*}
\end{small}\noindent
Here, $\vec{x}$ and $x$ are object terms, $P(\vec{x})$ is an arbitrary situation-suppressed SC formula, 
and $A\phi$ (i.e.\ \emph{over all paths} $\phi$) is a path quantifier. 
Also, $\bigcirc\phi$ means that $\phi$ holds {\em next} over a path while $\phi\;\mathcal{U}\;\psi$ stands for $\phi$ \emph{until} $\psi$. Finally, other logical connectives and quantifiers such as $\vee,\supset,\equiv,\exists$ and CTL$^*$ operators such as $\Diamond\phi$ (i.e.\ eventually $\phi$), $\phi\;\mathcal{B}\;\psi$ (i.e.\ $\phi$ before $\psi$), etc.\ are handled as the usual abbreviations. 
\par
Like $\mathit{now}$ in state formulae, path formulae $\phi$ can also contain an often-suppressed path placeholder $\mathit{path}$ in the place of the path terms. 
The function $\llbracket\cdot\rrbracket$ translates the above-defined formulae into formulae of the SC with paths. 
We write $\Phi\llbracket s\rrbracket$ (and $\phi\llbracket p\rrbracket$) to mean that state formula $\Phi$ (and path formula $\phi$) holds in situation $s$ (and over path $p$, respectively). 
See \cite{KL15} for how $\llbracket\cdot\rrbracket$ is defined. 
\par
We will use $\alpha$ and $\sigma$, possibly with decorations, to represent ground action and situation terms, respectively. Finally, we will use uppercase Latin letters for ground terms, and lowercase Latin letters for variables. 
%
%
\newline\noindent\textbf{Example}\indent
For our running example, 
we consider a couple of simple rescue drone agents $D_1$ and $D_2$ and their flight paths from one location to another. At anytime, an agent can be in any of the four locations $L_s,L_d, L_1,$ and $L_1'$. The geometry of the flight paths is captured using the non-fluent relation $\mathit{Route}(l,l')$, which states that there is a flight path from location $l$ to $l'$ (throughout, we assume that free variables are universally quantified from the outside):\footnote{We assume that all agents know all non-fluent facts.} 

\vspace{-4 mm}
\begin{small}
\begin{eqnarray*}
&&\hspace{-7 mm}(a).\;\mathit{Route}(l,l')\equiv[(l=L_s\wedge l'=L_1)\vee(l=L_s\wedge l'=L_1')\\
&&\hspace{14 mm}\mbox{}\vee(l=L_1\wedge l'=L_d)\vee(l=L_1'\wedge l'=L_d)].
\end{eqnarray*}
\end{small}

\vspace{-4 mm}
\noindent A controller agent $D_c$ is in charge of the overall mission and warns about potentially unsafe routes. Besides the $\mathit{inform}$ communicative action mentioned above, there are three additional actions in this domain. 
Action $\mathit{takeOff}(d,l)$ can be used by drone $d$ to take off from location $l$, $\mathit{flyTo}(d,l,l')$ takes $d$ from $l$ to $l'$, and $\mathit{land}(d,l)$ makes $d$ land at $l$. There are four fluents in this domain, 
$\mathit{At}(d,l,s)$,  $\mathit{Flying}(d,s)$, $\mathit{Vis}(d,l,s),$ and $\mathit{TStrom}(l,s)$, representing that $d$ is located at $l$ in situation $s$, that $d$ is flying in $s$, that $d$ has visited $l$ in $s$, and that there is an ongoing thunderstorm at $l$ in $s$.
\par
The action preconditions in this domain are as follows:

\vspace{-4 mm}
\begin{small}
\begin{eqnarray*}
&&\hspace{-7 mm}(b).\;\mathit{Poss}(\mathit{takeOff}(d,l),s)\equiv\mathit{At}(d,l,s)\land\neg\mathit{Flying}(d,s),\\
&&\hspace{-7 mm}(c).\;\mathit{Poss}(\mathit{flyTo}(d,l,l'),s)\equiv\mathit{At}(d,l,s)\land\mathit{Flying}(d,s)\\
&&\hspace{10 mm}\mbox{}\land\mathit{Route}(l,l')\land\neg\mathit{Know}(d,\mathit{TStrom}(l'),s),\\
&&\hspace{-7 mm}(d).\;\mathit{Poss}(\mathit{land}(d,l),s)\equiv\mathit{At}(d,l,s)\land\mathit{Flying}(d,s).
\end{eqnarray*}
\end{small}

\vspace{-4 mm}
\noindent Thus, e.g., $(c)$ states that a drone agent $d$ can fly from locations $l$ to $l'$ in situation $s$ iff it is located at $l$ in $s$, it is flying in $s$, there is a route from $l$ to $l'$, and it does not know that there is a thunderstorm at $l'$ in $s$. 
\par
Moreover, the SSA for the above fluents are as follows.

\vspace{-4 mm}
\begin{small}
\begin{eqnarray*}
&&\hspace{-7 mm}(e).\;\mathit{At}(d,l,do(a,s))\equiv[\exists l'.\;a=\mathit{flyTo}(d,l',l)\\
&&\hspace{22 mm}\mbox{}\lor(\mathit{At}(d,l,s)\land\neg\exists l'.\;a=\mathit{flyTo}(d,l,l'))],\\
&&\hspace{-7 mm}(f).\;\mathit{Flying}(d,do(a,s))\equiv[\exists l.\;a=\mathit{takeOff}(d,l)\\
&&\hspace{22 mm}\mbox{}\lor(\mathit{Flying}(d,s)\land\neg\exists l.\;a=\mathit{land}(d,l))],\\
%
&&\hspace{-7 mm}(g).\;\mathit{Vis}(d,l,do(a,s))\equiv
\exists l'.\;a=\mathit{flyTo}(d,l',l)\lor\mathit{Vis}(d,l,s),\\
&&\hspace{-7 mm}(h).\;\mathit{TStrom}(l,do(a,s))\equiv\mathit{TStrom}(l,s).
\end{eqnarray*}
\end{small}

\vspace{-4 mm}
\noindent Thus, e.g., Axiom $(e)$ states that $d$ is at location $l$ after executing action $a$ in situation $s$ (i.e.\ in $do(a,s)$) iff $a$ refers to $d$'s action of flying from some location $l'$ to $l$, or $d$ was already at $l$ in $s$ and $a$ is not its action of flying to a different location $l'$.
\par
Initially, drone $D_1$ is at location $L_s$, is not flying, and has only visited $L_s$, and it knows these facts. Moreover, it does not know that there is a storm at location $L_1$, but knows that there are no storms at $L_1'$ and $L_d$. There is indeed a thunderstorm at location $L_1$ and the controller agent $D_c$ knows this. Finally, $D_c$ does not know however that the other agents know this fact. These are captured using the following initial state axioms (note that $\mathit{Know}(d,\Phi(\mathit{now}),s)\supset\Phi[s]$):

\vspace{-4 mm}
\begin{small}
\begin{eqnarray*}
&&\hspace{-7 mm}(i).\;\mathit{Know}(D_1,\mathit{At}(D_1,L_s),S_0),\\ 
&&\hspace{-7 mm}(j).\;\mathit{Know}(D_1,\neg\mathit{Flying}(D_1),S_0),\\
&&\hspace{-7 mm}(k).\;\mathit{Know}(D_1,\forall l.\;\mathit{Vis}(D_1,l)\equiv l=L_s,S_0),\\
&&\hspace{-7 mm}(l).\;\neg\mathit{Know}(D_1,\mathit{TStrom}(L_1),S_0),\\
&&\hspace{-7 mm}(m).\;\mathit{Know}(D_1,\neg\mathit{TStrom}(L_1'),S_0),\\
&&\hspace{-7 mm}(n).\;\mathit{Know}(D_1,\neg\mathit{TStrom}(L_d),S_0),\\
&&\hspace{-7 mm}(o).\;\mathit{Know}(D_c,\mathit{TStrom}(L_1),S_0),\\
&&\hspace{-7 mm}(p).\;\forall d.\;d\neq D_c\supset\neg\mathit{Know}(D_c,\mathit{Know}(d,\mathit{TStrom}(L_1)),S_0).
\end{eqnarray*}
\end{small}
\vspace{-4 mm}
\section{Formalizing Goals and Intentions}
To model conative effects in the SC, we adopt the expressive formalization of prioritized goals (p-goals) and intentions proposed by KL \shortcite{KL10}. In this framework, each p-goal is specified by its own accessibility relation $G$. To deal with multiple agents, we modify KL's proposal by adding an agent argument for all goal-related predicates and relations; usually the first argument for this. Given agent $d$, a path $p$ is $G$-accessible at priority level $n$ in situation $s$, denoted by $G(d,p,n,s)$, iff the goal of $d$ at level $n$ is satisfied over $p$ and $p$ starts with a situation that has the same action history as $s$. The latter requirement ensures that the agent's p-goal-accessible paths reflect the actions that have been performed so far. A smaller $n$ represents higher priority, with $0$ being the highest priority level. Thus the set of p-goals are totally ordered according to priority. We say that $d$ has the p-goal that $\phi$ at level $n$ in situation $s$ iff $\phi$ holds over all paths that are $G$-accessible for $d$ at $n$ in $s$, i.e.\ $\mathit{PGoal}(d,\phi,n,s)\defi\forall p.\;G(d,p,n,s)\supset\phi\llbracket p\rrbracket.$
\par
%
We assume that a domain theory $\mathcal{D}$ for our framework also includes the domain-dependent initial goal axioms (see below) and the domain-independent axioms and definitions that appear throughout this paper. As KL, we allow the agent to have infinitely many goals, some of which can be left unspecified. For instance, assume that initially, our drone agent $D_1$ has the following two p-goals: $\phi_0=\Diamond\mathit{At}(D_1,L_d)$, i.e.\ that it is eventually at $L_d$, and $\phi_1=\mathit{Vis}(D_1,L_1)\;\mathcal{B}\;\mathit{Vis}(D_1,L_d)$, i.e.\ that it visits $L_1$ before it visits $L_d$, at level $0$ and $1$, respectively. Also, $D_c$ does not have any initial p-goals. Then the initial goal hierarchy of $D_1$ and $D_c$ can be specified using the following axioms:

\vspace{-4 mm}
\begin{small}
\begin{eqnarray*}
&&\hspace{-7 mm}(q).\;\mathit{Init}(s)\!\supset\!
((G(D_1,p,0,s)\equiv\mathit{Starts}(p,s')\!\wedge\!\mathit{Init}(s')\!\wedge\!\phi_0\llbracket p\rrbracket)\\
&&\hspace{9 mm}\mbox{}\wedge(G(D_1,p,1,s)\equiv\mathit{Starts}(p,s')\!\wedge\!\mathit{Init}(s')\!\wedge\!\phi_1\llbracket p\rrbracket),\\
%
&&\hspace{-7 mm}(r).\;
\mathit{Init}(s)\!\wedge\!n\geq 2\supset
(G(D_1,p,n,s)\equiv\mathit{Starts}(p,s')\!\wedge\!\mathit{Init}(s')),\\
&&\hspace{-7 mm}(s).\;
\mathit{Init}(s)\!\wedge\!n\geq 0\supset
(G(D_c,p,n,s)\equiv\mathit{Starts}(p,s')\!\wedge\!\mathit{Init}(s')).
\end{eqnarray*}
\end{small}

\vspace{-4 mm}
\noindent$(q)$ specifies the p-goals $\phi_0,\phi_1$ (from highest to lowest priority) of $D_1$ in the initial situations, and makes $G(D_1,p,n,s)$ true for every path $p$ that starts with an initial situation and over which $\phi_n$ holds, for $n=0,1$; each of them defines a set of initial goal paths for a given priority level, and must be consistent. $(r)$ makes $G(D_1,p,n,s)$ true for every path $p$ that starts with an initial situation for $n\geq 2$. Thus at levels $n\geq 2$, $D_1$ has the trivial p-goal that it be in an initial situation. The case for $D_c$ is similar. 
\par
Assume that $\mathcal{D}_{dr}$ denotes our theory for the drone domain. Then in our example, we can show the following:

\vspace{-4 mm}
\begin{small}
\begin{eqnarray*}
&&\hspace{-7 mm}\mathcal{D}_{dr}\models\mathit{PGoal}(D_1,\phi_n\!\wedge\!\mathit{Starts}(p,s)\!\wedge\!\mathit{Init}(s),n,S_0),\textup{ for }n<2,\\
&&\hspace{-7 mm}\mathcal{D}_{dr}\models\mathit{PGoal}(D_1,\mathit{Starts}(p,s)\wedge\mathit{Init}(s),n,S_0),\textup{ for any }n\geq 2 .
\end{eqnarray*}
\end{small}
\vspace{-4 mm}
\par
Since not all $G$-accessible paths are realistic in the sense that they start with a $K$-accessible situation, to filter the unrealistic paths out, KL defined {\em realistic} p-goal accessible paths:
\begin{eqnarray*}
G_R(d,p,n,s)\defi G(d,p,n,s)\wedge\mathit{Starts}(p,s')\wedge K(d,s',s).
\end{eqnarray*} 
$G_R$ prunes out the paths from $G$ that are known to be impossible, and since intentions are defined in terms of realistic p-goals, this ensures that these are realistic. 
\par
Using realistic p-goals-accessible paths, KL defined intentions as the realistic and maximal consistent prioritized intersection of the agent's goal hierarchy. 
%
%
First they specify all paths $p$ that are in this prioritized intersection $G_\cap(d,p,n,s)$:\footnote{\textbf{if}$ \phi$ \textbf{then} $\delta_1$ \textbf{else} $\delta_2$ is an abbreviation for $(\phi\supset\delta_1)\wedge(\neg\phi\supset\delta_2).$} 

\vspace{- 4mm}
\begin{small}
\begin{eqnarray*}
&&\hspace{-7 mm}G_\cap(d,p,n,s)\equiv
\textbf{\textup{if}}\;(n=0)\;\textbf{\textup{then}}\\
&&\hspace{18 mm}\textbf{\textup{if}}\;\exists p'.\;G_R(d,p',n,s)\;\textbf{\textup{then}}\;G_R(d,p,n,s)\\
&&\hspace{18 mm}\textbf{\textup{else}}\;\mathit{Starts}(p,s')\wedge K(d,s',s)\\
&&\hspace{15 mm}\textbf{\textup{else}}\\
&&\hspace{18 mm}\textbf{\textup{if}}\;\exists p'.(G_R(d,p',n,s)\wedge G_\cap(d,p',n-1,s))\\
&&\hspace{22 mm}\textbf{\textup{then}}\;(G_R(d,p,n,s)\wedge G_\cap(d,p,n-1,s))\\
&&\hspace{18 mm}\textbf{\textup{else}}\;G_\cap(d,p,n-1,s).
\end{eqnarray*}
\end{small}

\vspace{-4 mm}
Using this, they defined what it means for an agent to have an intention at some level $n$:\footnote{KL used the term ``chosen goals'' (C-Goals) for this.} 
$\mathit{Int}(d,\phi,n,s)\defi\forall p.\;G_\cap(d,p,n,s)\supset\phi\llbracket p\rrbracket$, 
i.e.\ an agent $d$ has the intention at level $n$ that $\phi$ in situation $s$ if $\phi$ holds over all paths that are in the prioritized intersection of $d$'s set of $G_R$-accessible paths up to level $n$ in $s$. 
Finally, intentions are defined in terms of intentions at $n$: 
$\mathit{Int}(d,\phi,s)\defi\forall n.\;\mathit{Int}(d,\phi,n,s)$, 
i.e.\ the agent $d$ has the intention that $\phi$ in $s$ if for any level $n$, $\phi$ is $d$'s intention at $n$ in $s$. 
\par
In our example, it can be shown that initially the $D_1$ has the intention that $\phi_0$ and that $\phi_1$: 
$\mathcal{D}_{dr}\models\mathit{Int}(D_1,\phi_0\wedge\phi_1,S_0)$.
\noindent\textbf{Goal Dynamics}\indent 
An agent's goals change when its knowledge changes as a result of the occurrence of an action, including exogenous events, or when it 
adopts or drops a goal. KL showed how this can be formalized by specifying how p-goals change. Intentions are then computed using realistic p-goals in every 
new situation as above. 
\par
Since for our example we only need to model cooperative agents that always respect the controller agent's requests, to simplify, we will modify KL's framework slightly by introducing a request communicative action and by getting rid of the actions for goal 
adoption and dropping. 
$req(d,d',\phi)$ can be used by an agent $d$ to request to adopt a p-goal $\phi$ to another agent $d'$. The APA for this is as follows:

\vspace{- 4mm}
\begin{small}
\begin{eqnarray*}
&&\hspace{-7 mm}\mathit{Poss}(\mathit{req}(d,d',\phi),s)\equiv\mbox{}\\
&&\hspace{-5 mm}\neg\mathit{Int}(d,\neg\exists s',p'.\;\mathit{Starts}(s')\wedge\mathit{Suffix}(p',do(\mathit{req}(d,d',\phi),s'))\\
&&\hspace{9 mm}\mbox{}\wedge\phi\llbracket p'\rrbracket,s)\land\neg\exists n.\;\mathit{PGoal}(d',\phi,n,s).
\end{eqnarray*}
\end{small}

\vspace{-4 mm}
\noindent That is, an agent $d$ can request another agent $d'$ to adopt the p-goal that $\phi$ if $d$ does not intend in $s$ that it is not the case that it executes the $\mathit{req}$ action next and $\phi$ holds afterwards, 
and $d'$ does not already have $\phi$ as its p-goal at some level $n$ in $s$.
\par
In the following, we specify the dynamics of p-goals by giving the SSA for $G$ and discuss each case, one at a time:

\vspace{-4 mm}
\begin{small}
\begin{eqnarray*}
&&\hspace{-5 mm}G(d,p,n,do(a,s))\equiv\\
&&\hspace{-3 mm}\forall d',\phi.\;(a\neq\mathit{req}(d',d,\phi)\wedge
\mathit{Progressed}(d,p,n,a,s))\\
&&\hspace{-3 mm}\mbox{}\vee\exists d',\phi.\;(a=\mathit{req}(d',d,\phi)\wedge\mathit{Requested}(d,p,n,a,s,\phi)).
\end{eqnarray*} 
\end{small}

\vspace{-4 mm}
\noindent The overall idea 
for this is as follows. First of all, to handle the occurrence of a non-request (i.e.\ a regular or a request not directed to $d$) action $a$, we progress all of $d$'s $G$-accessible paths to reflect the fact that $a$ has just happened; this is done using the $\mathit{Progressed}(d,p,n,a,s)$ construct, which replaces each of $d$'s $G$-accessible path $p'$ with starting situation $s'$, by its suffix $p$ provided that it starts with $do(a,s')$:

\vspace{-4 mm}
\begin{small}
\begin{eqnarray*}
&&\hspace{-7 mm}\mathit{Progressed}(d,p,n,a,s)\defi\\
&&\hspace{-4 mm}\exists p',s'.\;G(d,p',n,s)\wedge\mathit{Starts}(p',s')\wedge\mathit{Suffix}(p,p',do(a,s')).
\end{eqnarray*}
\end{small}

\vspace{-4 mm}
\noindent Any path over which the next action performed is not $a$ is eliminated from the respective $G$-accessibility level for $d$. 
\par
Secondly, to handle the request of a p-goal $\phi$ directed to $d$, we add a new proposition containing the p-goal to $d$'s goal hierarchy at the highest priority level by modifying the $G$-relation accordingly.\footnote{For simplicity, we assume that the requested goal is always adopted as the highest priority goal. Other sophisticated models, e.g.\ one where the requestee adopts the requested goal only if it is from a trusted source, it is consistent with its own set of core goals, and at just below these core goals, could have been modeled as easily.} 
The $G$-accessible paths for $d$ at level $0$ are the ones that share the same history with $do(a,s)$ and over which $\phi$ holds. The $G$-accessible paths for $d$ at all levels below $0$ are the ones that can be obtained by progressing the level immediately above it. Thus the agent $d$ acquires the p-goal that $\phi$ at the highest priority level $0$, and all the p-goals in $s$ are pushed down one level in the hierarchy. 

\vspace{-4 mm}
\begin{small}
\begin{eqnarray*}
&&\hspace{-7 mm}\mathit{Requested}(d,p,n,a,s,\phi)\defi\\
&&\hspace{-3 mm}\textbf{if}\;(n=0)\;\textbf{then}\;\\
&&\hspace{3 mm}\exists s'.\;\mathit{Starts}(p,s')
\wedge\mathit{SameHist}(s',do(a,s))\wedge\phi\llbracket p\rrbracket\\
&&\hspace{-3 mm}\textbf{else}\;\mathit{Progressed}(d,p,n-1,a,s).
\end{eqnarray*}
\end{small}

\vspace{-4 mm}
In our example, we can show that the agent $D_1$ will have the intention that $\Diamond\mathit{Vis}(D_1,L_1')$ after $D_1$ takes off from $L_s,$ $D_c$ informs $D_1$ that there is a thunderstorm at $L_1,$ and $D_c$ requests $D_1$ to eventually visit $L_1'$, starting in $S_0$, i.e.\ in situation $S_3=do([\mathit{takeOff}(D_1,L_s);\mathit{inform}(D_c,D_1,\mathit{TStrom}$ $(L_1));$ 
$\mathit{req}(D_c,D_1,\Diamond\mathit{Vis}(D_1,L_1'))],S_0);$ thus: 

\vspace{-2 mm}
\begin{small}
\[\mathcal{D}_{dr}\models\mathit{Int}(D_1,\Diamond\mathit{Vis}(D_1,L_1'),S_3).\]
\end{small}

\vspace{-4 mm}
\noindent But $D_1$ will not have the intention that $\phi_1$ as it has become impossible for $D_1$ to visit $L_1$ due to its knowledge of the thunderstorm at $L_1$, i.e.\ $\mathcal{D}_{dr}\models\neg\mathit{Int}(D_1,\phi_1,S_3)$.
\section{Handling Conative Effects}
Given a trace of events, \emph{actual achievement causes} are the events that are behind achieving an effect.\footnote{We do not conceptually distinguish between actions and events.} 
To formalize reasoning about epistemic effects, KL \shortcite{KL21} introduced the notion of \emph{epistemic dynamic formulae in the SC}. An effect in their framework is thus an epistemic dynamic formula. We will extend this notion to that of \emph{intentional dynamic formulae} $\varphi$ to deal with conative effects (see below). Given an effect $\varphi,$ the actual causes are defined relative to a {\em narrative} (variously known as a {\em scenario} or a {\em trace}) $s$. When $s$ is ground, the tuple $\langle\varphi,s\rangle$ is often called a {\em causal setting} \cite{BatusovS18}. Also, it is assumed that $s$ is executable, and $\varphi$ was false before the execution of the actions in $s$, but became true afterwards, i.e.\ 
$\mathcal{D}\models \mathit{executable}(s)\wedge\neg\varphi\lceil\mathit{root}(s)\rfloor\wedge\varphi\lceil s\rfloor$, where $\mathit{root}(s)\doteq\mathit{root}(s'),$ if $\exists a'.\;s=do(a',s'),$ and $\mathit{root}(s)=s,$ otherwise. Here $\varphi\lceil s\rfloor$  denotes the formula obtained from $\varphi$ by restoring the appropriate situation argument into all fluents in $\varphi$ (see Definition \ref{psiSAT}).
%
%
%
%
\par
Note that since all changes in the SC result from actions, the potential causes of an effect $\varphi$ are identified with a set of action terms occurring 
in $s$. However, since $s$ might include multiple occurrences of the same action, one also needs to identify the situations where these actions were executed. To deal with this, KL required that each situation is associated with a time-stamp. 
Since in the context of knowledge, we will have different $K$-accessible situations where an action occurs, using time-stamps provides a common reference/rigid designator for the action occurrence. 
The initial situations start at time 0 and each action increments the time-stamp by one. Thus, our 
theory includes the following axioms:

\vspace{-4 mm}
\begin{small}
\begin{eqnarray*}
&&\mathit{Init}(s)\supset\mathit{time}(s)=0,\\
&&\forall a,s,t.\;\mathit{time}(do(a,s))=t\equiv \mathit{time}(s)=t-1.
\end{eqnarray*}
\end{small}

\vspace{-4 mm}
\noindent With this, causes in this framework is a non-empty set of action-time-stamp pairs derived from the trace $s$ given $\varphi$. 
%
%
\par
We now introduce the notion of \emph{intentional dynamic formulae} (IF, henceforth):
\begin{definition}
Let 
$\vec{x}$, $\theta_a$, and $\vec{y}$ respectively range over object terms, action terms, and object and action terms. 
The class of \emph{situation-suppressed intentional dynamic formulae} $\varphi$ is defined inductively using the following grammar:

\vspace{-4 mm}
\begin{small}
\begin{eqnarray*}
&&\hspace{-5 mm}\varphi::=P(\vec{x}) \mid Poss(\theta_a)\mid\mathit{After}(\theta_a,\varphi)\mid\neg\varphi\mid\varphi_1\wedge\varphi_2\\
&&\hspace{10 mm}\mbox{}\mid\exists\vec{y}.\;\varphi\mid\mathit{Know}(\mathit{agt},\varphi)\mid\mathit{Int}(\mathit{agt},\psi).
\end{eqnarray*}
\end{small}
\end{definition}

\noindent That is, an IF can be a situation-suppressed fluent, a formula that says that some action $\theta_a$ is possible, a formula that some IF holds after some action has occurred, a formula that can built from other IF using the usual connectives, or a formula that the agent knows that some IF holds or intends to bring about some path formula $\psi$. 
Note that $\varphi$ can have quantification over object and action variables, but must not include quantification over situations or ordering over situations (i.e.\ $\sqsubset$)  
or arbitrary $K$ or $G$-relations, i.e.\ those that do not come from the expansion of $\mathit{Know}/\mathit{Int}$ . 
We will use $\varphi$ for IF. 
\par
Note that the argument of $\mathit{Int}$ in the above inductive definition is a path formula $\psi$. Thus to allow for IF in the context of $\mathit{Int}$, we need to redefine state formulae $\Phi$ to include IF $\varphi$:

\vspace{-2 mm}
\begin{small}
\[\Phi::=P(\vec{x})\mid A \phi\mid\Phi\wedge\Phi\mid\neg\Phi\mid\forall x.\;\Phi\mid\varphi.\]
\end{small}

\vspace{-2 mm}
We define $\varphi\lceil\cdot\rfloor$ as follows:
\begin{definition}\label{psiSAT}
\begin{small}
\begin{eqnarray*}
&&  \hspace{-7 mm}\varphi\lceil s\rfloor\defi
    \begin{cases}      
      P(\vec{x},s) & \textup{ if }\varphi\textup{ is }P(\vec{x})\\
      \mathit{Poss}(\theta_a,s) & \textup{ if }\varphi\textup{ is }\mathit{Poss}(\theta_a)\\
      \varphi'\lceil do(\theta_a,s)\rfloor & \textup{ if }\varphi\textup{ is }\mathit{After}(\theta_a,\varphi')\\
      \neg(\varphi'\lceil s\rfloor) & \textup{ if }\varphi\textup{ is }(\neg\varphi')\\
      \varphi_1\lceil s\rfloor\wedge\varphi_2\lceil s\rfloor & \textup{ if }\varphi\textup{ is }(\varphi_1\wedge\varphi_2)\\
      \exists\vec{y}.\;(\varphi'\lceil s\rfloor) & \textup{ if }\varphi\textup{ is }(\exists\vec{y}.\;\varphi')\\
      \forall s'.\;K(d,s',s)\supset(\varphi'\lceil s'\rfloor) & \textup{ if }\varphi\textup{ is }\mathit{Know}(d,\varphi')\\
      \forall n.\;\mathit{Int}(d,\psi,n,s) & \textup{ if }\varphi\textup{ is }\mathit{Int}(d,\psi).\\
    \end{cases}      
\end{eqnarray*}
\end{small}
\end{definition}
\par
We will now present the definition of causes in the SC. The idea behind how causes are computed is as follows. Given an effect $\varphi$ and scenario $s$, if some action of the action sequence in $s$ triggers the formula $\varphi$ to change its truth value from false to true relative to $\mathcal{D}$, and if there are no actions in $s$ after it that change the value of $\varphi$ back to false, then this action is an actual cause of achieving $\varphi$ in $s$. Such causes are referred to as {\em primary} causes:\footnote{This is a slightly generalized definition than that of \cite{KL21}, where the authors used $S_0$ instead of $\mathit{root}(s)$.}
\begin{definition}[Primary Cause]\label{PCause}
\begin{small}
\begin{eqnarray*}
&&\hspace{-10 mm}\mathit{CausesDirectly}(a,t,\varphi,s)\defi\mbox{}\\
&&\hspace{3 mm}\exists s_a.\;\mathit{time}(s_a)=t\wedge(\mathit{root}(s)<do(a,s_a)\leq s)\\
&&\hspace{3 mm}\mbox{}\wedge\neg\varphi\lceil s_a\rfloor\wedge\forall s'.(do(a,s_a)\leq s'\leq s\supset\varphi\lceil s'\rfloor).
\end{eqnarray*}
\end{small}
\end{definition}
\noindent That is, $a$ executed at time $t$ is the \emph{primary cause} of effect $\varphi$ in situation $s$ 
iff $a$ was executed in a situation with time-stamp $t$ in scenario $s$, $a$ caused $\varphi$ to change its truth value to true, and no subsequent actions on the way to $s$ falsified $\varphi$. 
\par
Now, note that a (primary) cause $a$ might have been non-executable initially. Also, $a$ might have only brought about the effect conditionally and this context condition might have been false initially. Thus earlier actions on the trace that contributed to the preconditions and the context conditions of a cause must be considered as a cause as well. The following definition captures both primary and indirect causes.\footnote{In this, we need to quantify over situation-suppressed IF. Thus we must encode such formulae as terms and formalize their relationship to the associated SC formulae. This is tedious but can be done essentially along the lines of \cite{GiacomoLL00}. We assume that we have such an encoding and use formulae as terms directly.} 
\begin{definition}[Actual Cause \cite{KL21}]\label{ACause}
\begin{small}
\begin{eqnarray*}
&&\hspace{-7 mm}\mathit{Causes}(a,t,\varphi,s)\defi\\
&&\hspace{-7 mm}\forall P.[
\forall a,t,s,\varphi.(\mathit{CausesDirectly}(a,t,\varphi,s)\supset P(a,t,\varphi,s))\wedge\mbox{}\\
&&\hspace{-5 mm}\forall a,t,s,\varphi.( \exists a'\!,t'\!,s'\!.(\mathit{CausesDirectly}(a'\!,t'\!,\varphi,s)
    \land \mathit{time}(s')\!=\!t'\\
&&\hspace{12 mm}\mbox{} \land s'<s \land
P(a,t,[\mathit{Poss}(a')\wedge\mathit{After}(a',\varphi)],s'))
\\
&&\hspace{23 mm}\mbox{}\supset P(a,t,\varphi,s))\\
&&\hspace{-3 mm}]\supset P(a,t,\varphi,s).
\end{eqnarray*}
\end{small}
\end{definition}
\noindent Thus, $\mathit{Causes}$ is defined to be the least relation $P$ such that if $a$ executed at time $t$ directly causes $\varphi$ in scenario $s$ then $(a,t,\varphi,s)$ is in $P$, and if $a'$ executed at $t'$ is a direct cause of $\varphi$ in $s$, the time-stamp of $s'$ is $t'$, $s'<s$, and $(a,t,[\mathit{Poss}(a')\wedge\mathit{After}(a',\varphi)],s')$ is in $P$ (i.e.\ $a$ executed at $t$ is a direct or indirect cause of $[\mathit{Poss}(a')$ $\mbox{}\wedge\mathit{After}(a',\varphi)]$ in $s'$), then $(a,t,\varphi,s)$ is in $P$. Here the effect $[\mathit{Poss}(a')\wedge\mathit{After}(a',\varphi)]$ requires $a'$ to be executable and $\varphi$ to hold after $a'$.
%
%
\par%
With these simple modifications, the framework is now capable of dealing with conative effects. To see this, consider the following scenario $\sigma$ in our example. 
%
$\sigma=do([\mathit{takeOff}(D_1,L_s);
\mathit{inform}(D_c,D_1,\mathit{TStrom}(L_1));$
$\mathit{req}(D_c,D_1,\Diamond\mathit{Vis}(D_1,L_1'));
\mathit{inform}(D_c,\!D_2,\!\mathit{TStrom}(L_1));$
$\mathit{req}(D_c,D_2,\Diamond\mathit{Vis}(D_1,L_1'));
\mathit{flyTo}(D_1,L_s,L_1');
\mathit{flyTo}(D_1,$ $L_1',L_d)],S_0).$ 
There are 7 actions in this scenario. For convenience, we will use $\vec{\alpha_i}$ to denote the first $i$ actions in this trace, and so $do([\vec{\alpha_5}],S_0)$ is the situation obtained from executing the first 5 actions starting in $S_0$. 
Now assume that we want to reason about the causes of the effect $\varphi_1=\mathit{Int}(D_1,\Diamond\mathit{Vis}(D_1,L_1'))$ in scenario $\sigma_1=do([\vec{\alpha_5}],S_0)).$ Then we can show that:

\vspace{-4 mm}
\begin{small}
\[\mathcal{D}_{dr}\models\mathit{Causes}(\mathit{req}(D_c,D_1,\Diamond\mathit{Vis}(D_1,L_1')),2,\varphi_1,\sigma_1),\]
\end{small}

\vspace{-4 mm}
\noindent i.e.\ as expected, $D_c$'s request to $D_1$ to eventually visit $L_1'$ that was executed at time 2 is the cause of $D_1$'s intention that $\Diamond\mathit{Vis}(D_1,L_1').$
\section{Reasoning about Agent Behaviour}
We are now ready to formalize reasoning about agent behaviour via causation. Just like causes, an explanation in our framework is also modeled using an action-time-stamp pair $(a,t)$. Agent behaviour, on the other hand, is captured using a situation $s$ and relative to an observation $\varphi$. For this, we use the predicate $\mathit{Explains}(a,t,\varphi,s)$, which means that the action $a$ executed at time $t$ explains the behaviour of the agents captured in situation $s$ relative to the observation $\varphi$. For example, $\mathit{Explains}(\alpha,\tau,\varphi_2,\sigma)$ states that the behaviour of drones as modeled by situation/scenario $\sigma$ relative to the effect that $\varphi_2=\mathit{Vis}(D_1,L_1')$ can be explained by action $\alpha$ executed at time $\tau$ (see below for the values of $\alpha$ and $\tau$). Thus, $(\alpha,\tau)$ explains why the drone $D_1$ visited the location $L_1'$. Note that, just as in the case for achievement causation, we assume here that $\neg\varphi\lceil\mathit{root}(s)\rfloor\wedge\varphi\lceil s\rfloor$. 
\par
While explaining agent behaviour through direct causation is reasonable, it may not always be insightful. For instance, we can show that agent behaviour in $\sigma$ w.r.t.\ visiting $L_1'$ can be explained by its action $\mathit{flyTo}(D_1,L_s,L_1')$ executed at time $5$. However, this is obvious and is hardly useful. A deeper level of explanation requires analyzing the mental states of the involved agents. 
\par
To further explain agent behaviour, we will use an intention recognition system, which for this paper is considered to be a black-box module. We use the predicate $\mathit{RRInt}(d,\phi,a,t,s)$ to denote that agent $d$ is recognized to have the relevant intention that $\phi$ in situation $s$ w.r.t.\ the action $a$ executed at time $t$. For instance, $\mathit{RRInt}(D_1,\Diamond\mathit{Vis}(D_1,L_1'),\mathit{flyTo}(D_1,L_s,L_1'),5,\sigma)$ says that in scenario $\sigma$, agent $D_1$ is recognized to have the intention that $\Diamond\mathit{Vis}(D_1,L_1')$ for executing the action $\mathit{flyTo}(D_1,L_s,L_1')$ at time $5$. With this, we can further explain agent behaviour via the root-cause analysis of its intentions behind performing actions. In our example, since $D_1$ flew to $L_1'$ due to its intention that $\Diamond\mathit{Vis}(D_1,L_1'),$ it is reasonable to explain agent behaviour via the causes of having this intention. This will reveal that $D_1$ had this intention due to $D_c$'s request, and thus agent behaviour w.r.t.\ $D_1$ visiting $L_1'$ can explained by this request action. 
\par
We now give the definition for $\mathit{Explains}$:
\begin{definition}\label{Explains}
\begin{small}
\begin{eqnarray*}
&&\hspace{-7 mm}\mathit{Explains}(a,t,\varphi,s)\defi\neg\varphi\lceil\mathit{root}(s)\rfloor\wedge\varphi\lceil s\rfloor\wedge\\
&&\hspace{-3.5 mm}[\mathit{Causes}(a,t,\varphi,s)\vee\mbox{}\\
&&\hspace{3 mm}(\exists a',t',d',s',\psi.\;\mathit{Explains}(a',t',\varphi,s)\wedge\mathit{agent}(a')=d'\\
&&\hspace{5 mm}\mbox{}\wedge\mathit{RRInt}(d',\psi,a',t',s)\wedge s'<s\wedge\mathit{time}(s')=t'\\
&&\hspace{5 mm}\mbox{}\wedge\neg\mathit{Int}(d',\psi,\mathit{root}(s'))\wedge\mathit{Int}(d',\psi,s')\\
&&\hspace{5 mm}\mbox{}\wedge\mathit{Causes}(a,t,\mathit{Int}(d',\psi),s'))].
\end{eqnarray*}
\end{small}
\end{definition}
\noindent Thus, agent behaviour relative to the observation that $\varphi$ in scenario $s$ can be explained by the action $a$ executed at time $t$ iff $a$ at $t$ is a cause of $\varphi$ in $s$; or some other action $a'$ executed at time $t'$ explains $\varphi$ in $s$, the agent of $a'$ is $d'$, $d'$ is recognized to have the intention that $\psi$ behind performing $a'$ at $t'$ in $s$, and $a$ executed at $t$ was the cause of this intention in $s'$. Here $\mathit{agent}(a)$ denotes the agent of the action $a$; it can be specified as usual by an axiom that returns the agent of $a$, usually the first argument of $a$, i.e.\ $\mathit{agent}(a(d,\vec{x}))=d.$ Also, $s'$ is the situation where $a'$ was executed. Finally, the two requirements that the effect is false before the execution of the actions in the scenario and became true afterwards, i.e.\ $\neg\varphi\lceil\mathit{root}(s)\rfloor\wedge\varphi\lceil s\rfloor$ and $\neg\mathit{Int}(d',\psi,\mathit{root}(s'))\wedge\mathit{Int}(d',\psi,s')$, are needed to ensure that the causes actually exist.
\par
Returning to our example, we now formally state the two explanations that we mentioned above and give the values for $\alpha$ and $\tau$. First, we can show that:

\vspace{-3 mm}
\begin{small}
\[\mathcal{D}_{dr}\models\mathit{Explains}(\mathit{flyTo}(D_1,L_s,L_1'),5,\varphi_2,\sigma).\]
\end{small}

\vspace{-4 mm}
\noindent But perhaps more interestingly, we can show that:

\vspace{-4 mm}
\begin{small}
\begin{eqnarray*}
&&\hspace{-7 mm}\mathcal{D}_{dr}\cup\{\mathit{RRInt}(D_1,\Diamond\mathit{Vis}(D_1,L_1'),\mathit{flyTo}(D_1,L_s,L_1'),5,\sigma)\}\models\\
&&\mathit{Explains}(\mathit{req}(D_c,D_1,\Diamond\mathit{Vis}(D_1,L_1')),2,\varphi_2,\sigma).
\end{eqnarray*}
\end{small}

\vspace{-4 mm}
It is important to  note that the scenario $s$ in Definition \ref{Explains} may and will often include the actions of multiple agents, and thus explanation of agent behaviour may trigger the analysis of the mental states of multiple agents. For example, given a different scenario, recognizing the intention behind the controller agent $D_c$'s request to $D_1$ and analyzing this intention can in turn expose the causes behind its actions, e.g.\ due to its prior commitments to safety, etc. As such, the analysis performed here is truly multiagent in nature. Also, although our example only involves single-action causes and we do not consider epistemic effects, as discussed above, the framework does support secondary causes and causal knowledge dynamics; see \cite{KL21} for concrete examples. 
\section{Conclusion}
In this paper, we formalized causal reasoning about motivations. Using this, we offer a novel take on explainable AI that is grounded in theory of mind: agent behaviour in our framework can be explained via the causal analysis of observed effects, which as we show can trigger the analysis of their mental states. This paper reports our ongoing work. 
Understanding the properties of our proposal and how it relates to previous work in this area is what we plan to investigate next. 

\section*{Acknowledgments}
We acknowledge the support of the Natural Sciences and Engineering Research Council of Canada (NSERC), [funding reference number RGPIN-2022-03433].
\mbox{}\newline\mbox{}\newline
\noindent Cette recherche a été financée par le Conseil de recherches en sciences naturelles et en génie du Canada (CRSNG), [numéro de référence RGPIN-2022-03433].


\bibliographystyle{named}
\bibliography{explain-xai}

\end{document}